%% file: IGARSS2026LaTeXTemplate.tex
\documentclass[conference,a4paper]{IEEEtran}
\IEEEoverridecommandlockouts

\usepackage[hidelinks]{hyperref}
\usepackage[cmex10]{amsmath}
\usepackage{amssymb,amsfonts}
\usepackage[ruled,vlined]{algorithm2e}
\usepackage{graphicx}
\graphicspath{{Figures/PDF/}{Figures/PNG/}}

\usepackage{booktabs}
\usepackage{siunitx}
\usepackage[numbers,compress]{natbib}
\usepackage{texnames}
\usepackage{bm,bbm}
\usepackage{orcidlink}
\usepackage{booktabs,multirow}
\usepackage{booktabs,multirow}
\usepackage{svg}
\usepackage{amsmath}
\usepackage{float}
\usepackage{stfloats}
\usepackage{hyperref}
\usepackage{etoolbox}
\hypersetup{
    colorlinks=true,
    citecolor=green,
    linkcolor=blue,
    urlcolor=blue
}

\usepackage{eso-pic}

\usepackage[font=footnotesize]{caption}
\makeatletter
\def\@maketitle{%
\newpage
\bgroup
\par\vskip\IEEEtitletopspace\vskip\IEEEtitletopspaceextra\centering%
\vskip0.2em%
{\normalfont\fontsize{19}{21}\selectfont\@title\par}%
\vskip1.0em\par%
{\@IEEEspecialpapernotice\mbox{}\vskip\@IEEEauthorblockconfadjspace%
\mbox{}\hfill\begin{@IEEEauthorhalign}\@author\end{@IEEEauthorhalign}\hfill\mbox{}\par}%
\relax
\par\addvspace{0.5\baselineskip}\egroup}
\makeatother
\begin{document}

\title{MITIGATING LONG-TAIL BIAS VIA PROMPT-CONTROLLED\\
DIFFUSION AUGMENTATION}

\author{\scriptsize	\IEEEauthorblockN{Buddhi Wijenayake\orcidlink{0009-0001-2624-0251}}
	\IEEEauthorblockA{\textit{University of Peradeniya}\\
		Peradeniya, Sri Lanka\\
		e19445@eng.pdn.ac.lk}
	\and
	\IEEEauthorblockN{Nichula Wasalathilake\orcidlink{0009-0003-5662-009X}}
	\IEEEauthorblockA{\textit{University of Peradeniya}\\
		Peradeniya, Sri Lanka\\
		e20425@eng.pdn.ac.lk}\\
         
	\IEEEauthorblockN{Parakrama Ekanayake\orcidlink{0000-0002-5639-8105}}
	\IEEEauthorblockA{\textit{University of Peradeniya}\\
		Peradeniya, Sri Lanka\\
		mpbe@eng.pdn.ac.lk}
	\and
	\IEEEauthorblockN{Roshan Godaliyadda\orcidlink{0000-0002-3495-481X}}
	\IEEEauthorblockA{\textit{University of Peradeniya}\\
		Peradeniya, Sri Lanka\\
		roshang@eng.pdn.ac.lk}\\
        
        \IEEEauthorblockN{Vishal M. Patel\orcidlink{0000-0002-5239-692X}}
	\IEEEauthorblockA{\textit{Johns Hopkins University}\\
		Baltimore, Maryland, USA\\
		vpatel36@jhu.edu}
    \and 
    \IEEEauthorblockN{Vijitha Herath\orcidlink{0000-0001-5083-7632}}
	\IEEEauthorblockA{\textit{University of Peradeniya}\\
		Peradeniya, Sri Lanka\\
		vijitha@eng.pdn.ac.lk}
	}
\maketitle
\AddToShipoutPictureFG*{%
  \AtPageUpperLeft{%
    \hspace{0.8cm}\raisebox{-1.2cm}{%
      \parbox{0.7\textwidth}{\textit{Accepted for publication at IEEE IGARSS 2026}}%
    }%
  }%
}
\vspace*{-2\baselineskip}
\begin{abstract}
	Long-tailed class imbalance remains a fundamental obstacle in semantic segmentation of high-resolution remote-sensing imagery, where dominant classes shape learned representations and rare classes are systematically under-segmented. This challenge becomes more acute in cross-domain settings such as LoveDA, which exhibits an explicit Urban/Rural split with substantial appearance differences and inconsistent class-frequency statistics across domains. We propose a prompt-controlled diffusion augmentation framework that generates paired label-image samples with explicit control over semantic composition and domain, enabling targeted enrichment of underrepresented classes rather than indiscriminate dataset expansion. A domain-aware, masked, ratio-conditioned discrete diffusion model first synthesizes layouts that satisfy class-ratio targets while preserving realistic spatial co-occurrence, and a ControlNet-guided diffusion model then renders photorealistic, domain-consistent images from these layouts. When mixed with real data, the resulting synthetic pairs improve multiple segmentation backbones, especially on minority classes and under domain shift, showing that better downstream segmentation comes from adding the right samples in the right proportions.
\end{abstract}
\input{Figure_Texs/dataset}
\vspace*{-1.7\baselineskip}
{Source codes, pretrained Models and synthetic datasets are available at \href{https://buddhi19.github.io/SyntheticGen}{\texttt{buddhi19.github.io/SyntheticGen}}.}

\input{Chapters/introduction}
\input{Chapters/methodology}

\input{Tables/results_main}

\input{Chapters/resultsanddiscussions}

\input{Chapters/conclusion}

\small
\bibliographystyle{IEEEtranN}
\bibliography{references}

\end{document}

%% file: Figure_Texs/dataset.tex
\begin{figure*}[b]
    \centering
    \includegraphics[width=1\linewidth]{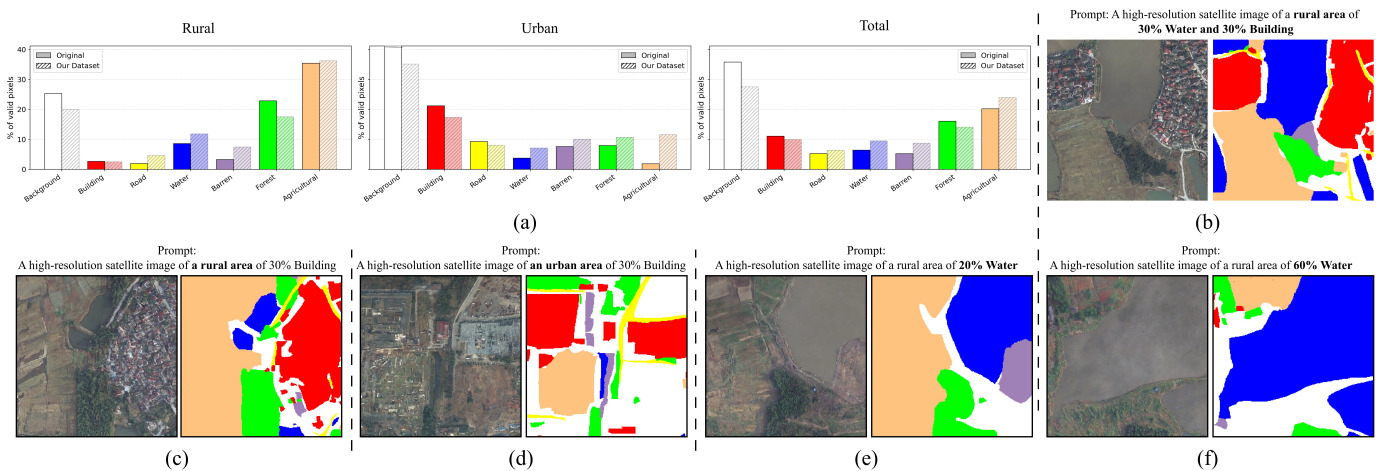}
    \caption{Dataset balancing and prompt-controllable synthesis on LoveDA. (a) Pixel-frequency distributions for Rural, Urban, and the combined training set, comparing the original data (solid) against our augmented dataset (hatched). (b--f) Representative synthesized image--label pairs generated under explicit domain (Urban/Rural) and class-ratio constraints, illustrating controllable diffusion for both domain-consistent appearance and targeted semantic proportions.}
    \label{fig:dataset}
\end{figure*}

%% file: Chapters/introduction.tex
\section{Introduction}
\label{sec:introduction}

Semantic segmentation of high-resolution remote sensing imagery supports key geospatial applications such as urban mapping, land-cover monitoring, and environmental assessment \cite{Yuan2020DLEnvironmentalRS,Ma2019DLRemoteSensing}. However, performance in realistic settings is often limited by severe pixel-level class imbalance where a few dominant categories occupy most pixels, while minority classes appear sparsely, creating a long-tailed training signal that biases learning toward frequent classes and degrades rare-class recognition.

This challenge is amplified in LoveDA \cite{Wang2021LoveDA}, which is explicitly organized into Urban and Rural domains with distinct scene structure, appearance, and inconsistent class distributions across domains (Fig.~\ref{fig:dataset}(a)). Several semantically important categories occupy only a small fraction of pixels, and the tail classes differ between Urban and Rural splits \cite{Wang2021LoveDA}. As a result, models must jointly address within-domain long-tail imbalance and cross-domain shift, an interaction that standard supervised training often struggles to resolve \cite{Wang2021LoveDA}.

Common remedies include class re-weighting \cite{Cui2019ClassBalancedLoss}, focal-style objectives \cite{Lin2017FocalLoss}, online hard example mining \cite{Shrivastava2016OHEM}, resampling, and geometric/photometric augmentation \cite{Cubuk2019AutoAugment}. While these strategies can stabilize optimization, they largely preserve the underlying pixel-frequency statistics and cannot reliably increase exposure to rare, spatially localized, context-dependent classes. Moreover, LoveDA’s domain-dependent imbalance makes naive re-weighting prone to domain-specific overfitting or head-class gains without consistent tail improvements across both splits \cite{Wang2021LoveDA}.

Generative augmentation offers a complementary alternative by synthesizing additional labeled data rather than only reshaping the loss. Diffusion models provide strong fidelity and diversity \cite{Ho2020DDPM,Rombach2022LDM}, and recent Earth observation studies have shown their potential for satellite image generation and layout-conditioned synthesis \cite{Sebaq2024RSDiff,Baghirli2023SatDM}. For long-tailed segmentation, however, realism alone is insufficient; what matters is controllable generation of the right samples. If synthesis merely follows the empirical training distribution, rare classes remain rare and domain imbalance can persist \cite{Kim2025SampleEfficientGDA,Niemeijer2025UncertaintyControlNet,Yu2023DistributionShiftInversion,Qin2023ClassBalancingDiffusion}.

To address these limitations, we propose a prompt-controlled generative augmentation framework that explicitly conditions synthesis on domain and class-ratio targets. The framework generates domain-consistent label-image pairs that increase minority class exposure while preserving realistic remote sensing structure, thereby improving downstream segmentation not by indiscriminately expanding the dataset, but by adding the right data in the right proportions.

%% file: Chapters/methodology.tex
\section{Methodology}
\input{Figure_Texs/stageA}
The proposed architecture is a two-stage, domain-aware generative augmentation pipeline that synthesizes paired label--image samples with controllable semantic composition. In the first Stage (Stage A), a ratio and domain-conditioned discrete diffusion model (D3PM) \cite{Austin2021DiscreteDDPM} generates semantic layouts that match desired class-ratio targets within a specified domain. In the second stage (Stage B), a fine-tuned Stable Diffusion network translates these layouts into photorealistic remote-sensing images while preserving spatial structure and domain appearance. The resulting synthetic corpus is mixed with the real training set using a controlled sampling protocol to train the state-of-the-art segmentation models. Subsequent sections present each contribution in detail.

\subsection{Stage A: Ratio and Domain conditioned Discrete Layout Diffusion}

Diffusion models define a forward noising process and learn a reverse denoising process to generate samples. While standard DDPMs operate in continuous spaces, Discrete Denoising Diffusion Probabilistic Models (D3PMs) extend this framework to categorical variables, making them well-suited for discrete structures such as semantic label maps \cite{Austin2021DiscreteDDPM}.

\input{Figure_Texs/stageB}
As seen in figure \ref{fig:stageA}, given a semantic map $x_0^{label} \in \{1,\dots,K\}^{H_\ell \times W_\ell}$ with $K$ land-cover classes, we define its class-ratio vector as,
\begin{equation}
\begin{aligned}
r(x_0^{label} )_k &= \frac{1}{|\Omega|}\sum_{(i,j)\in \Omega}\mathbbm{1}[x_0^{label} (i,j)=k],\\
\text{with}\quad&\sum_{k=1}^K r_k = 1,
\end{aligned}
\label{eq:ratio}
\end{equation}
where $\Omega$ excludes ignored pixels. The semantic map will be downsampled and one-hot encoded to form $x_0 \in K\times 256\times 256$.

\paragraph{Forward corruption}
Stage~A uses a D3PM forward Markov chain \cite{Austin2021DiscreteDDPM} given by,
\begin{equation}
q(x_{1:T}\mid x_0)=\prod_{t=1}^{T} q(x_t\mid x_{t-1}),
\end{equation}
which is parameterized by a categorical transition matrix $Q_t\in\mathbb{R}^{K\times K}$. The scheduler samples the corrupted one-hot layout $x_t \sim q(x_t\mid x_0,t)$\cite{Austin2021DiscreteDDPM}.

\paragraph{Conditioning}\label{sec:cond}
We condition the denoiser on (i) a ratio target and (ii) a domain label $d\in\{\text{Urban},\text{Rural}\}$. To support partial control, we randomly mask ratio constraints during training using $m\in\{0,1\}^{K}$, where $m_k=1$ indicates that class $k$ is constrained. Inside the embedding converter shown in figure \ref{fig:stageA}, a lightweight ratio projector maps the masked ratio input to a conditioning vector $e_r\in\mathbb{R}^{d_e}$ aligned with the diffusion time-embedding dimension. A learnable domain embedding $e_d\in\mathbb{R}^{d_e}$ is added to obtain the final conditioning embedding as,
$e = e_r + \alpha\, e_d$
where $\alpha$ is a learnable scalar controlling the domain contribution.

\paragraph{Reverse model}
A UNet denoiser $f_\theta$ takes the corrupted one-hot layout $x_t$, timestep $t$, and conditioning $e$, and outputs per-pixel class logits as,
\begin{equation}
\ell_\theta(x_t,t,e) = f_\theta(x_t,t;e)\in\mathbb{R}^{K\times H_\ell \times W_\ell}.
\end{equation}

\input{Figure_Texs/inference}

\paragraph{Training objective}
Given logits $\ell_\theta$, we compute per-pixel class probabilities
$p_\theta=\mathrm{softmax}(\ell_\theta)$ and estimate the global class-ratio vector $\hat r_k$ by averaging these
probabilities over valid layout pixels similar to equation \ref{eq:ratio}.

We then apply a two-weight
ratio-matching loss that prioritizes constrained classes while softly regularizing unconstrained ones, which can be given as,
\begin{equation}
\mathcal{L}_{\text{ratio}} \;=\;
\| m \odot (\hat r - r) \|_2^2
+0.1\, \| (1-m)\odot (\hat r - r) \|_2^2.
\end{equation}
The first term enforces the requested ratios, while the $0.1$-weighted term encourages the model to complete
the remaining composition using learned co-occurrence statistics rather than arbitrary allocation.

Following \cite{Austin2021DiscreteDDPM}, we train the discrete diffusion model by minimizing the negative
variational lower bound $\mathcal{L}_{\text{VLB}}$ with an auxiliary denoising cross-entropy term
$\mathcal{L}_{\text{CE}}$ for stabilization, and add the masked ratio constraint to obtain the stage A loss function,
\begin{equation}
\mathcal{L}_A \;=\; \mathcal{L}_{\text{VLB}} \;+\; 0.5\,\mathcal{L}_{\text{CE}}
\;+\mathcal{L}_{\text{ratio}}.
\end{equation}

\subsection{Stage B: Layout-guided Image Synthesis with Ratio and Domain-aware ControlNet}

As seen in figure \ref{fig:placeholder}, Stage~B synthesizes a photorealistic remote-sensing image conditioned on (i) a semantic layout, and (ii) a domain-aware prompt. We build on latent diffusion models (LDMs) \cite{Rombach2022LDM} and ControlNet \cite{Zhang2023ControlNet}, and use FiLM gating \cite{Perez2018FiLM} to modulate ControlNet residual features.

\paragraph{Latent diffusion}
Given an RGB remote-sensing image $I\in\mathbb{R}^{3\times H\times W}$, a VAE encoder maps it to a latent representation $z_0\in\mathbb{R}^{4\times \frac{H}{8}\times \frac{W}{8}}$. A noise scheduler corrupts this latent as,
\begin{equation}
\begin{aligned}
    z_t &= \alpha_t z_0 + \sigma_t \epsilon,\\
    \text{with}\quad\epsilon \sim \mathcal{N}(0,I),&\quad t\sim \mathcal{U}\{0,\dots,T-1\},
\end{aligned}
\end{equation}
where $(\alpha_t,\sigma_t)$ follow the diffusion schedule \cite{Rombach2022LDM}. The denoising model predicts the injected noise $\epsilon$ from $(z_t,t)$ under multi-source conditioning.

\paragraph{Conditioning signals}
A domain prompt (e.g., \emph{``a high-resolution satellite image of an urban area of 10\% water''} or \emph{``a high-resolution satellite image of a rural area of 30\% building, 5\% agriculture and 3\% forest''}) is encoded by a CLIP text encoder \cite{Radford2021CLIP}to obtain token embeddings $c_{\text{text}}$. These embeddings condition both the ControlNet U-Net and the fine-tuned Stable Diffusion U-Net via cross-attention \cite{Rombach2022LDM}.

The semantic map $x_0^{label}$ is converted into a $K$-channel one-hot tensor and fed to ControlNet as the conditioning input.

\paragraph{Gated ControlNet residual injection}
Motivated by prior results showing that feature-wise affine modulation preserves semantic-layout conditioning in image synthesis \cite{Park2019SPADE}, we adapt a ControlNet that takes the noisy latent, timestep, text embeddings, and one-hot layout conditioning, and outputs
multi-scale residual features as,
\begin{equation}
\{\Delta^{(b)}\},\,\Delta^{(\mathrm{mid})}
=\mathrm{ControlNet}\!\left(z_t,t,c_{\text{text}},x_0\right),
\end{equation}
where $\Delta^{(b)}$ denotes residuals at the downsampling blocks and $\Delta^{(\mathrm{mid})}$ the mid-block residual
\cite{Zhang2023ControlNet}. 
 Before injecting these residuals into the main denoising U-Net, we apply a
FiLM-style feature-wise affine gate \cite{Perez2018FiLM} to regulate residual strength. This can be expressed as,
\begin{equation}
\begin{aligned}
\tilde{\Delta}^{(b)} &= \gamma^{(b)} \odot \Delta^{(b)} + \beta^{(b)},\\
\tilde{\Delta}^{(\mathrm{mid})} &= \gamma^{(\mathrm{mid})} \odot \Delta^{(\mathrm{mid})} + \beta^{(\mathrm{mid})},
\end{aligned}
\end{equation}
where $(\gamma,\beta)$ are learnable scale and bias parameters and $\odot$ represents element wise multiplication. 

\paragraph{Training objective}
Stage~B is trained with the standard noise-prediction objective used in DDPM/LDMs \cite{Ho2020DDPM, Rombach2022LDM}.
For each training iteration, the denoiser predicts
\begin{equation}
\hat\epsilon \;=\; \epsilon_\theta\!\left(z_t, t, c_{\text{text}};\{\tilde{\Delta}^{(b)}\},\tilde{\Delta}^{(\mathrm{mid})}\right),
\end{equation}
and we minimize the mean-squared error to the injected noise,
\begin{equation}
\mathcal{L}_B
\;=\;
\mathbb{E}_{I\sim\mathcal{D},\, t,\, \epsilon}
\left[
\left\lVert \epsilon - \hat{\epsilon} \right\rVert_2^2
\right].
\end{equation}

\subsection{Inference}
As given in the figure \ref{fig:infer}, at the inference, we first sample a semantic layout using Stage~A. From the input prompt, we extract the domain label $d$ and the
user-specified ratio targets $r$ and form the conditioning
embedding $e$. The discrete reverse chain is initialized from categorical noise $x_{T_1}$ and run for $T_1$ steps. At each
step, the U-Net predicts logits $\ell_\theta$, which parameterize the reverse transition distribution
$p_\theta(x_{t-1}\mid x_t,e)$, and we sample $x_{t-1}\sim p_\theta(x_{t-1}\mid x_t,e)$. After the final step, we obtain the
generated one-hot layout $\hat{x}_0$, which is upsampled to the conditioning resolution required by Stage~B.

Stage~B then synthesizes an image conditioned on the generated layout and the same prompt. We initialize the latent from
Gaussian noise $z_{T_2}\sim\mathcal{N}(0,\mathbf{I})$ and iteratively denoise for $T_2$ steps using a latent diffusion sampler,
conditioned on (i) the CLIP prompt embeddings $c_{\text{text}}$ and (ii) the upsampled layout $\hat{x}_0$
provided to ControlNet for spatial guidance. The final latent is decoded by the VAE decoder to obtain the synthetic satellite
image $\hat{I}$.

\subsection{Synthetic Dataset Construction and Downstream Segmentation}
\label{sec:synth_downstream}
Starting from running pixel-frequency statistics over real and accepted synthetic masks, we use a greedy enrichment strategy that repeatedly selects the most underrepresented non-background class in each domain and proposes ratio constraints to upweight it. Candidate layouts are accepted only if realized ratios satisfy the domain-specific constraints within a tolerance. We created 894 Rural and 1106 Urban additional samples (altogether 2000 pairs of images and labels), which are mixed with the original LoveDA training split. We then train five representative segmentation models, U-Net~\cite{Ronneberger2015UNet}, PSPNet~\cite{Zhao2017PSPNet}, FactSeg~\cite{Ma2022FactSeg}, HRNet~\cite{Wang2021HRNet}, and AerialFormer~\cite{Hanyu2024AerialFormer}, using an identical recipe on (i) real-only and (ii) real+synthetic data, and evaluate on the official LoveDA validation split. Results are reported in terms of mIoU and per-class IoU, emphasizing minority-class gains and cross-domain robustness.

%% file: Figure_Texs/stageA.tex
\begin{figure}
    \centering
    \includegraphics[width=1\linewidth]{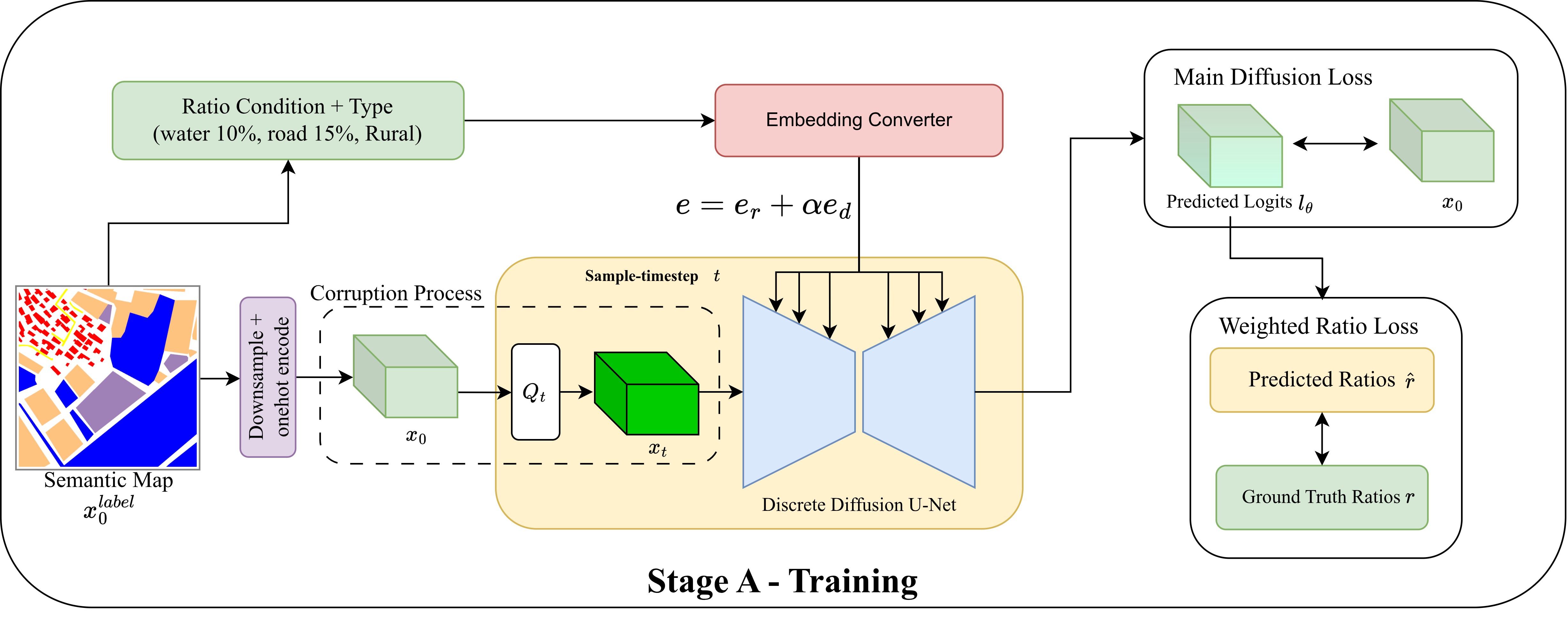}
\caption{Stage~A: domain and ratio conditioned discrete diffusion (D3PM) for semantic layout generation. A U-Net denoiser predicts categorical logits from a noisy label map conditioned on a masked class-ratio target and Urban/Rural domain embedding.}

    \label{fig:stageA}
\end{figure}

%% file: Figure_Texs/stageB.tex
\begin{figure}
    \centering
    \includegraphics[width=1\linewidth]{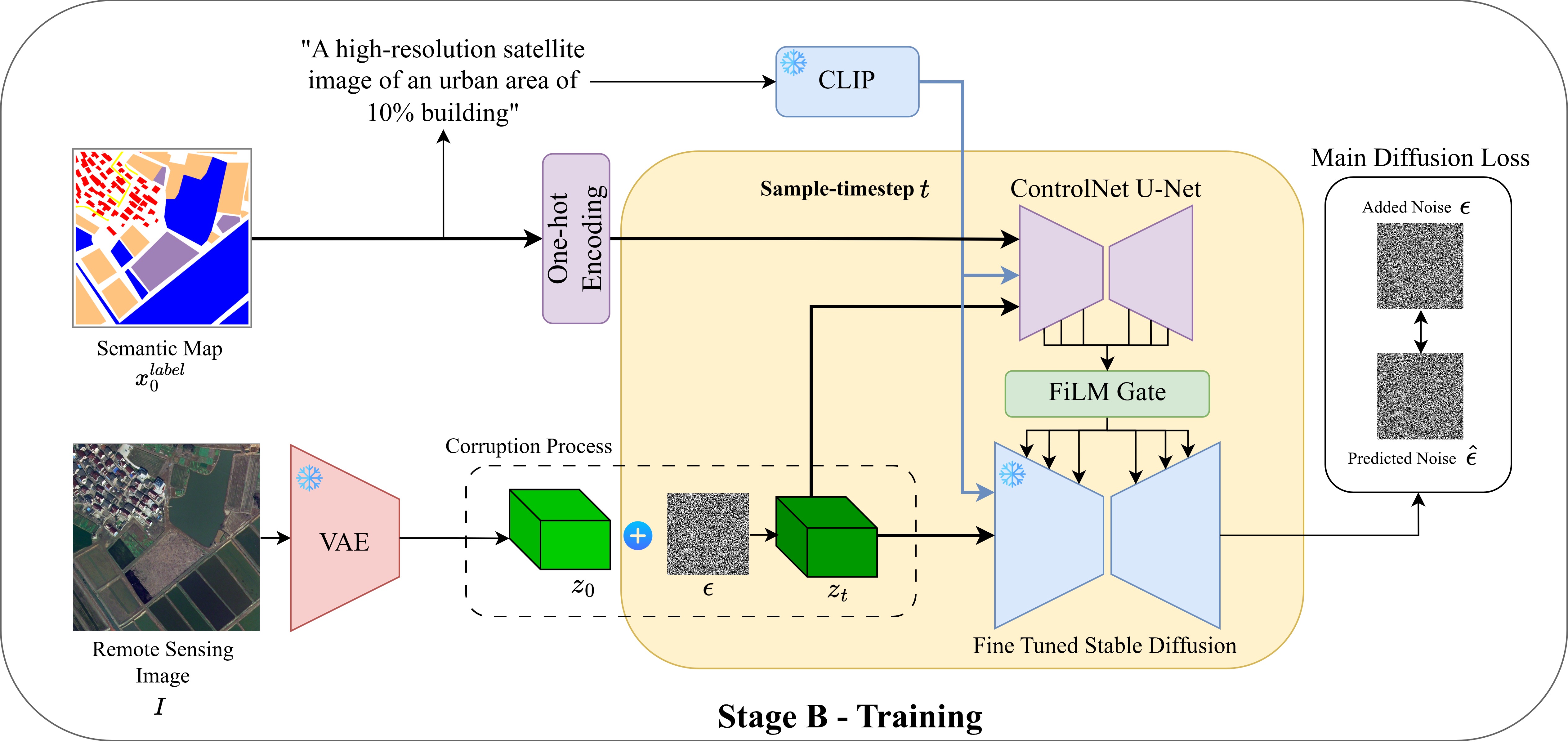}
    \caption{Stage~B: layout-guided latent diffusion for image synthesis. A Stable Diffusion U-Net is guided by ControlNet features from the layout, with FiLM-gated residual injection and a domain/ratio prompt for domain-consistent appearance.}
    \label{fig:placeholder}
\end{figure}

%% file: Figure_Texs/inference.tex
\begin{figure}
    \centering
    \includegraphics[width=1\linewidth]{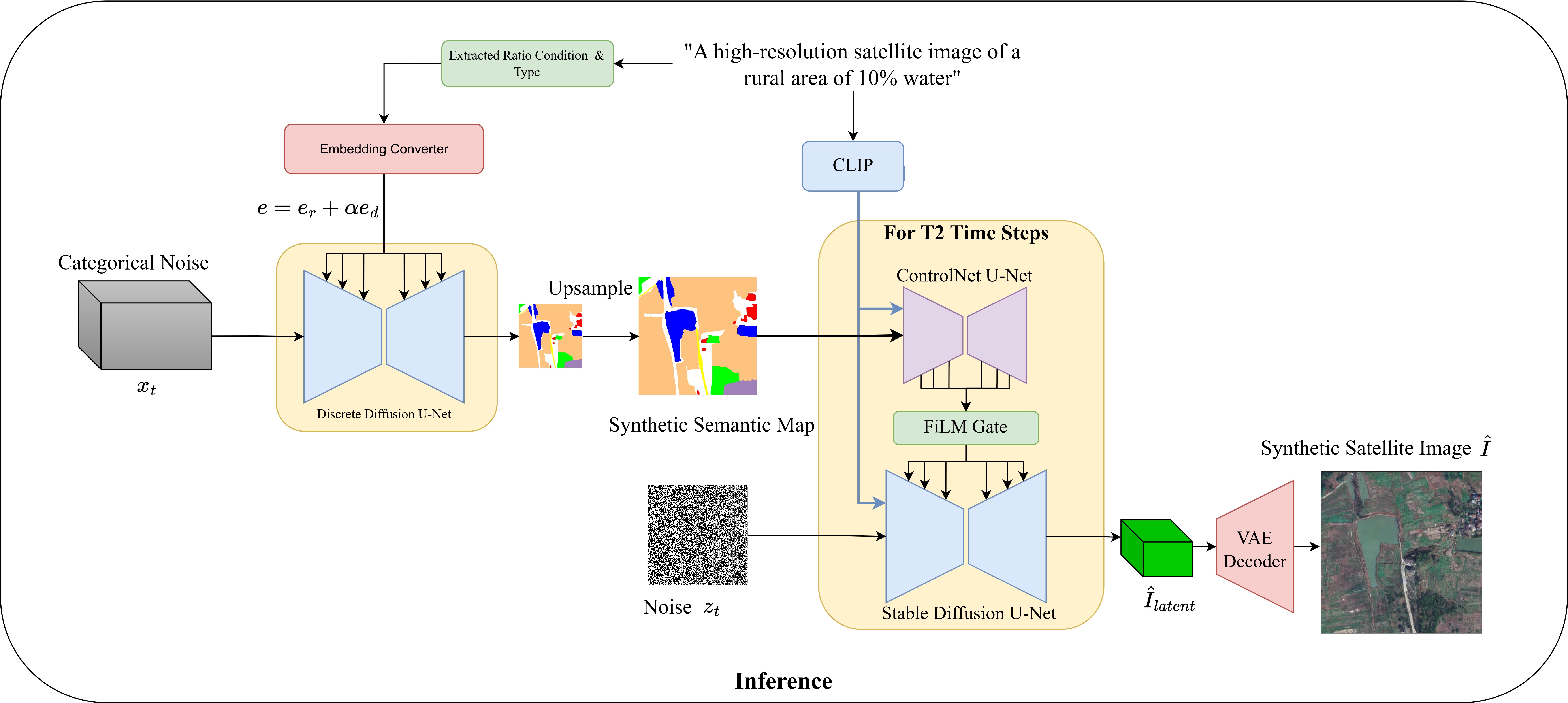}
    \caption{Prompt-controlled inference pipeline. The prompt is parsed into domain and ratio targets, a layout is sampled with Stage~A, and a photorealistic image is generated with Stage~B using the sampled layout as spatial guidance.}
    \label{fig:infer}
\end{figure}

%% file: Tables/results_main.tex

\begin{table*}[t]
\centering
\caption{Combined downstream segmentation results with Original vs.\ Original+Synthetic training.
Colors/bold indicate the gain of Original+Synthetic over the corresponding Original value for each cell:
\textcolor{red}{red} ($\geq$+10), \textcolor{blue}{blue} (+5 to +10), \textbf{bold} (0 to +5).}
\label{tab:combined_synth_results}
\renewcommand{\arraystretch}{1.08}
\setlength{\tabcolsep}{3.5pt}
\scriptsize
\begin{tabular}{llcccccccccc}
\toprule
\multirow{2}{*}{Model} & \multirow{2}{*}{Training Dataset} & \multicolumn{7}{c}{IoU (\%)} & \multirow{2}{*}{mIoU} & \multirow{2}{*}{mF1} & \multirow{2}{*}{OA} \\
\cmidrule(lr){3-9}
& & Background & Building & Road & Water & Barren & Forest & Agriculture & & & \\
\midrule

\multicolumn{12}{l}{\textbf{(A) In-domain evaluation}}\\
\multirow{2}{*}{U-Net~\cite{Ronneberger2015UNet}}
& Orig.      & 45.91 & 53.74 & 45.27 & 58.10 & 17.13 & 35.13 & 23.11 & 39.77 & 58.22 & 55.35 \\
& Orig.+Syn. & \textcolor{blue}{54.79} & \textcolor{blue}{63.10} & \textcolor{blue}{52.68} & \textcolor{red}{68.66} & \textcolor{red}{27.43} & \textcolor{blue}{40.20} & \textcolor{red}{52.64} & \textcolor{red}{51.36} & \textcolor{red}{70.24} & \textcolor{red}{66.85} \\
\midrule
\multirow{2}{*}{PSPNet~\cite{Zhao2017PSPNet}}
& Orig.      & 51.31 & 57.23 & 41.14 & 50.45 & 19.22 & 36.27 & 45.45 & 43.01 & 64.20 & 59.14 \\
& Orig.+Syn. & \textbf{52.53} & \textbf{61.95} & \textcolor{blue}{48.76} & \textcolor{blue}{59.54} & \textbf{21.41} & \textbf{40.00} & \textbf{47.93} & \textbf{47.45} & \textbf{67.11} & \textbf{63.26} \\
\midrule
\multirow{2}{*}{FactSeg~\cite{Ma2022FactSeg}}
& Orig.      & 50.66 & 55.09 & 48.67 & 63.40 & 26.28 & 36.75 & 41.22 & 46.01 & 64.83 & 62.16 \\
& Orig.+Syn. & \textbf{52.79} & \textcolor{blue}{62.40} & \textbf{53.03} & \textbf{66.96} & 24.80 & \textcolor{blue}{42.98} & \textcolor{blue}{46.71} & \textbf{49.95} & \textbf{65.57} & \textcolor{blue}{68.11} \\
\midrule
\multirow{2}{*}{HRNet~\cite{Wang2021HRNet}}
& Orig.      & 50.31 & 59.15 & 45.53 & 62.97 & 31.08 & 39.71 & 43.21 & 47.43 & 65.55 & 63.68 \\
& Orig.+Syn. & \textcolor{blue}{55.44} & \textbf{61.21} & \textcolor{red}{55.77} & \textcolor{blue}{71.99} & 29.04 & 38.14 & \textcolor{red}{59.49} & \textcolor{blue}{53.01} & \textcolor{blue}{71.87} & \textbf{68.20} \\
\midrule
\multirow{2}{*}{AerialFormer~\cite{Hanyu2024AerialFormer}}
& Orig.      & 54.40 & 66.09 & 56.04 & 69.44 & 28.99 & 45.48 & 50.96 & 53.06 & 68.40 & 65.57 \\
& Orig.+Syn. & \textbf{55.26} & \textbf{67.55} & \textbf{56.23} & 67.13 & \textbf{32.50} & \textbf{45.92} & \textbf{55.21} & \textbf{54.26} & \textbf{69.61} & \textbf{68.35} \\
\midrule

\multicolumn{12}{l}{\textbf{(B) Domain generalization}}\\
\multicolumn{12}{l}{\textbf{Urban $\rightarrow$ Rural}}\\
\multirow{2}{*}{FactSeg~\cite{Ma2022FactSeg}}
& Orig.      & 53.10 & 41.24 & 29.58 & 22.80 & 12.60 & 10.80 & 42.38 & 30.36 & 44.55 & 61.44 \\
& Orig.+Syn. & \textbf{53.68} & \textbf{42.95} & \textcolor{blue}{34.97} & \textcolor{red}{33.81} & \textbf{12.72} & \textcolor{red}{22.19} & \textcolor{blue}{48.35} & \textcolor{blue}{35.52} & \textcolor{blue}{50.87} & \textbf{63.48} \\
\midrule
\multirow{2}{*}{HRNet~\cite{Wang2021HRNet}}
& Orig.      & 51.09 & 46.50 & 27.07 & 25.57 & 15.45 &  8.06 & 27.99 & 28.82 & 42.84 & 57.29 \\
& Orig.+Syn. & \textbf{55.25} & 45.20 & \textcolor{blue}{35.42} & \textcolor{red}{37.11} & \textbf{16.56} & \textbf{8.40} & \textcolor{red}{45.45} & \textcolor{blue}{34.77} & \textcolor{blue}{49.04} & \textcolor{blue}{64.20} \\
\midrule

\multicolumn{12}{l}{\textbf{Rural $\rightarrow$ Urban}}\\
\multirow{2}{*}{FactSeg~\cite{Ma2022FactSeg}}
& Orig.      & 36.92 & 42.54 & 18.88 & 61.36 & 34.72 & 38.85 & 46.59 & 39.98 & 56.07 & 58.38 \\
& Orig.+Syn. & \textbf{39.07} & \textcolor{red}{57.00} & \textcolor{red}{49.17} & \textbf{63.86} & \textcolor{blue}{42.25} & \textcolor{red}{52.14} & \textbf{49.65} & \textcolor{red}{50.45} & \textcolor{red}{66.71} & \textcolor{blue}{65.16} \\
\midrule
\multirow{2}{*}{HRNet~\cite{Wang2021HRNet}}
& Orig.      & 38.78 & 41.31 & 30.24 & 68.65 & 39.29 & 42.95 & 46.43 & 43.95 & 60.30 & 60.51 \\
& Orig.+Syn. & \textbf{40.54} & \textcolor{red}{58.17} & \textcolor{red}{50.63} & \textbf{70.69} & \textcolor{blue}{46.04} & \textcolor{red}{53.46} & \textcolor{red}{57.02} & \textcolor{blue}{53.79} & \textcolor{blue}{69.52} & \textcolor{blue}{68.53} \\
\bottomrule
\end{tabular}
\end{table*}

%% file: Chapters/resultsanddiscussions.tex
\section{Results and Discussion}
Figure \ref{fig:dataset} shows that the original LoveDA split is strongly long-tailed and domain-dependent, with minority categories receiving very limited pixel supervision. The mixed distribution has a higher exposure to minority classes without breaking domain realism, consistent with the controlled examples in Figure \ref{fig:dataset}(b--f).

Table \ref{tab:combined_synth_results} shows that adding ratio-controlled synthetic pairs consistently improves segmentation across backbones, with the largest gains concentrated on minority and mid-tail classes rather than only the head classes. In-domain, mIoU increases for all models, with particularly strong improvements in agriculture, road, and water, indicating that synthesis mainly contributes context diversity for underrepresented semantics while respecting Urban/Rural style constraints. In domain generalization, synthesis has also improved transfer in both directions, suggesting reduced reliance on source-domain co-occurrence shortcuts. 

%% file: Chapters/conclusion.tex
\section{Conclusion}
\label{sec:conclusion}

We presented a prompt-controlled diffusion augmentation framework for semantic segmentation under long-tailed class imbalance and domain shift in remote-sensing imagery. By conditioning generation on domain identity and class-ratio targets, the method synthesizes label-image pairs that selectively enrich underrepresented classes while preserving domain-consistent realism. Experiments on LoveDA show consistent gains across segmentation backbones, especially for minority classes, together with reduced Urban/Rural domain gap. 

More broadly, this work shows that augmentation is more effective when it is controllable across the factors that drive downstream failure, such as class composition and domain. The key value of the proposed framework is therefore not simply to add more data, but to add the right data in the right proportions. Future work will extend this idea to more datasets, tasks, and stronger forms of controllable generative augmentation.